\pdfoutput=1

\documentclass[11pt]{article}

\usepackage[]{ACL2023}

\usepackage{times}
\usepackage{latexsym}

\usepackage[T1]{fontenc}

\usepackage[utf8]{inputenc}
\usepackage[hang,flushmargin]{footmisc}

\usepackage{microtype}

\usepackage{inconsolata}

\usepackage{booktabs}       
\usepackage{url}            
\usepackage{amsfonts}       
\usepackage{amsmath}
\usepackage{graphicx}
\usepackage{multirow}
\usepackage{pifont}
\usepackage{natbib}

\title{Multilingual E5 Text Embeddings: A Technical Report}

\author{Liang Wang,~Nan Yang,~Xiaolong Huang,
~Linjun Yang,~Rangan Majumder,~Furu Wei\\
Microsoft Corporation \\
\{wangliang,nanya,fuwei\}@microsoft.com
}

\begin{document}
\maketitle
\begin{abstract}
This technical report presents the training methodology and evaluation results of the open-source multilingual E5 text embedding models,
released in mid-2023.
Three embedding models of different sizes (small / base / large) are provided,
offering a balance between the inference efficiency and embedding quality.
The training procedure adheres to the English E5 model recipe,
involving contrastive pre-training on 1 billion multilingual text pairs,
followed by fine-tuning on a combination of labeled datasets.
Additionally,
we introduce a new instruction-tuned embedding model,
whose performance is on par with state-of-the-art, English-only models of similar sizes.
Information regarding the model release can be found at \url{https://github.com/microsoft/unilm/tree/master/e5}.
\end{abstract}

\section{Introduction}
Text embeddings serve as fundamental components in information retrieval systems
and retrieval-augmented language models.
Despite their significance,
most existing embedding models are trained exclusively on English text~\citep{Reimers2019SentenceBERTSE,Ni2021LargeDE,ni2022sentence},
thereby limiting their applicability in multilingual contexts.

In this technical report,
we present the multilingual E5 text embedding models (\emph{mE5-\{small / base / large\}}),
which extend the English E5 models~\citep{wang2022text}.
The training procedure adheres to the original two-stage methodology:
weakly-supervised contrastive pre-training on billions of text pairs,
followed by supervised fine-tuning on small quantity of high-quality labeled data.
We also release an instruction-tuned embedding model~\footnote{Here instructions refer to the natural language descriptions of the embedding tasks.}
\emph{mE5-large-instruct} by utilizing the synthetic data from ~\citet{wang2023improving}.
Instructions can better inform embedding models about the task at hand,
thereby enhancing the quality of the embeddings.

For model evaluation,
we first demonstrate that our multilingual embeddings exhibit competitive performance
on the English portion of the MTEB benchmark~\citep{muennighoff2023mteb},
and the instruction-tuned variant even surpasses strong English-only models of comparable sizes.
To showcase the multilingual capability of our models,
we also assess their performance on the MIRACL multilingual retrieval benchmark~\citep{Zhang2023MIRACLAM}
across $16$ languages and on Bitext mining~\citep{zweigenbaum2018overview,Artetxe2019MassivelyMS}
in over $100$ languages.

\section{Training Methodology}

\begin{table}[ht]
\centering
\scalebox{0.95}{\begin{tabular}{lc}
\hline
          & \# Sampled \\ \hline
Wikipedia &  150M    \\
mC4  &  160M  \\
Multilingual CC News   &  160M \\
NLLB  &   160M  \\
Reddit  &  160M   \\
S2ORC  &   50M   \\
Stackexchange   &  50M \\
xP3  &  80M   \\
Misc. SBERT Data   &   10M     \\ \hline
Total          &    $\sim$1B        \\ \hline
\end{tabular}}
\caption{Data mixture for contrastive pre-training.}
\label{tab:pretrain_data}
\end{table}

\noindent
\textbf{Weakly-supervised Contrastive Pre-training }
In the first stage,
we continually pre-train our model on a diverse mixture of multilingual text pairs
obtained from various sources as listed in Table~\ref{tab:pretrain_data}.
The models are trained with a large batch size $32k$ for a total of $30k$ steps,
which approximately goes over $\sim1$ billion text pairs.
We employ the standard InfoNCE contrastive loss with only in-batch negatives,
while other hyperparameters remain consistent with the English E5 models~\citep{wang2022text}.

\begin{table}[ht]
\centering
\scalebox{0.95}{\begin{tabular}{lc}
\hline
          & \# Sampled \\ \hline
MS-MARCO Passage &  500k       \\
MS-MARCO Document & 70k \\
NQ, TriviaQA, SQuAD &  220k     \\
NLI  &   275k   \\
ELI5   &  100k      \\
NLLB & 100k \\
DuReader Retrieval  &  86k      \\
Fever &  70k  \\
HotpotQA   &  70k  \\
Quora Duplicate Questions   & 15k   \\
Mr. TyDi   &   50k    \\
MIRACL   &   40k   \\ \hline
Total          &  $\sim$1.6M       \\ \hline
\end{tabular}}
\caption{Data mixture for supervised fine-tuning.}
\label{tab:sup_finetune_data}
\end{table}

\noindent
\textbf{Supervised Fine-tuning }
In the second stage,
we fine-tune the models from the previous stage on a combination of high-quality labeled datasets.
In addition to in-batch negatives,
we also incorporate mined hard negatives and knowledge distillation from a cross-encoder model
to further enhance the embedding quality.
For the \emph{mE5-\{small / base / large\}} models released in mid-2023,
we employ the data mixture shown in Table~\ref{tab:sup_finetune_data}.

For the \emph{mE5-large-instruct} model,
we adopt the data mixture from ~\citet{wang2023improving},
which includes additional $500k$ synthetic data generated by GPT-3.5/4~\citep{OpenAI2023GPT4TR}.
This new mixture encompasses $150k$ unique instructions and covers $93$ languages.
We re-use the instruction templates from ~\citet{wang2023improving}
for both the training and evaluation of this instruction-tuned model.

\section{Experimental Results}

\begin{table}[ht]
\centering
\scalebox{0.95}{\begin{tabular}{lc}
\hline
 & MTEB (56 datasets) \\ \hline
LaBSE & 45.2 \\
Cohere$_\text{multilingual-v3}$ & 64.0 \\
BGE$_\text{large-en-v1.5}$ & 64.2  \\ \hline
mE5$_\text{small}$ &  57.9 \\
mE5$_\text{base}$ & 59.5  \\
mE5$_\text{large}$ & 61.5  \\
mE5$_\text{large-instruct}$ & \textbf{64.4} \\ \hline
\end{tabular}}
\caption{Results on the English portion of the MTEB benchmark.
LaBSE~\citep{feng2022language} is exclusively trained on translation pairs.
Limited information is available regarding the training data and model size are available for Cohere$_\text{multilingual-v3}$(\url{https://txt.cohere.com/introducing-embed-v3/}).
BGE$_\text{large-en-v1.5}$~\citep{xiao2023c} is an English-only model.
Full results are in Appendix Table ~\ref{tab:app_full_results}.}
\label{tab:english_benchmark}
\end{table}

\noindent
\textbf{English Text Embedding Benchmark }
Multilingual embedding models should be able to perform well on English tasks as well.
In Table~\ref{tab:english_benchmark},
we compare our models with other multilingual and English-only models on the MTEB benchmark~\citep{muennighoff2023mteb}.
Our best mE5 model surpasses the previous state-of-the-art multilingual model Cohere$_\text{multilingual-v3}$,
by $0.4$ points and outperforms a strong English-only model,
BGE$_\text{large-en-v1.5}$, by $0.2$ points.
While smaller models demonstrate inferior performance,
their faster inference and reduced storage costs render them advantageous for numerous applications.
\newline

\begin{table}[ht]
\centering
\scalebox{0.95}{\begin{tabular}{lcc}
\hline
 & nDCG@10 & R@100 \\ \hline
BM25 & 39.3 & 78.7  \\
mDPR & 41.5 & 78.8 \\ \hline
mE5$_\text{small}$ &  60.8  & 92.4 \\
mE5$_\text{base}$ &  62.3 &  93.1 \\
mE5$_\text{large}$ & \textbf{66.5} &  94.3 \\
mE5$_\text{large-instruct}$ & 65.7 & \textbf{94.6} \\ \hline
\end{tabular}}
\caption{Multilingual retrieval on the development set of the MIRACL benchmark.
Numbers are averaged over $16$ languages.}
\label{tab:miracl}
\end{table}

\noindent
\textbf{Multilingual Retrieval }
We evaluate the multilingual retrieval capability of our models using the MIRACL benchmark~\citep{Zhang2023MIRACLAM}.
As shown in Table~\ref{tab:miracl},
mE5 models significantly outperform mDPR,
which has been fine-tuned on the MIRACL training set,
in both nDCG@10 and recall metrics.
Detailed results on individual languages are provided in Appendix Table~\ref{tab:app_full_miracl_ndcg}.

\begin{table}[ht]
\centering
\scalebox{0.95}{\begin{tabular}{lcc}
\hline
 & \begin{tabular}[c]{@{}c@{}}BUCC 2018\\ 4 langs\end{tabular} & \begin{tabular}[c]{@{}c@{}}Tatoeba\\ 112 langs\end{tabular} \\ \hline
mContriever$_\text{msmarco}$ & 93.7 & 37.7 \\
LaBSE & 98.8 & 81.1 \\ \hline
mE5$_\text{small}$ & 93.2  & 64.2 \\
mE5$_\text{base}$ & 98.1 &  68.1 \\
mE5$_\text{large}$ & 98.6 &  75.7 \\
mE5$_\text{large-instruct}$ & \textbf{99.0} & \textbf{83.8} \\ \hline
\end{tabular}}
\caption{Bitext mining results.
mContriever~\citep{Izacard2021TowardsUD} numbers are run by ourselves based on the released checkpoint.}
\label{tab:bitext}
\end{table}

\noindent
\textbf{Bitext Mining }
is a cross-lingual similarity search task that requires the matching of two sentences with little lexical overlap.
As demonstrated in Table ~\ref{tab:bitext},
mE5 models exhibit competitive performance across a broad range of languages,
both high-resource and low-resource.
Notably,
the mE5$_\text{large-instruct}$ model surpasses the performance of LaBSE,
a model specifically designed for bitext mining,
due to the expanded language coverage afforded by the synthetic data~\citep{wang2023improving}.

\section{Conclusion}
In this brief technical report,
we introduce multilingual E5 text embedding models that are trained with a multi-stage pipeline.
By making the model weights publicly available,
practitioners can leverage these models for information retrieval, semantic similarity, and clustering tasks
across a diverse range of languages.

\bibliography{anthology,custom}
\bibliographystyle{acl_natbib}

\appendix

\begin{table*}[ht]
\centering
\scalebox{0.9}{\begin{tabular}{lcccccccc}
\hline
& \multicolumn{4}{c}{nDCG@10} & \multicolumn{4}{c}{R@100} \\ \hline
 & mE5$_\text{small}$ & mE5$_\text{base}$ & mE5$_\text{large}$ & E5$_\text{large-instruct}$ & mE5$_\text{small}$ & mE5$_\text{base}$ & mE5$_\text{large}$ & E5$_\text{large-instruct}$ \\ \hline
ar & 71.4 & 71.6 & 76.0 & 76.8 & 96.2 & 95.9 & 97.3 & 97.5 \\
bn & 68.2 & 70.2 & 75.9 & 73.9 & 97.4 & 96.6 & 98.2 & 98.2 \\
en & 48.0 & 51.2 & 52.9 & 51.5 & 85.3 & 86.4 & 87.6 & 88.2 \\
es & 51.2 & 51.5 & 52.9 & 53.7 & 87.6 & 88.6 & 89.1 & 89.3  \\
fa & 53.3 & 57.4 & 59.0 & 59.4 & 90.4 & 91.2 & 92.9 & 92.9  \\
fi & 73.3 & 74.4 & 77.8 & 77.3 & 96.3 & 96.9 & 98.1 & 97.9  \\
fr & 47.6 & 49.7 & 54.5 & 53.7 & 89.5 & 90.0 & 90.6 & 91.7  \\
hi & 55.2 & 58.4 & 62.0 & 60.3 & 91.0 & 92.6 & 93.9 & 94.1  \\
id & 50.7 & 51.1 & 52.9 & 52.1 & 86.2 & 87.4 & 87.9 &  88.4 \\
ja & 63.6 & 64.7 & 70.6 & 69.0 & 95.2 & 96.0 & 97.1 & 96.9  \\
ko & 61.2 & 62.2 & 66.5 & 65.3 & 92.0 & 91.6 & 93.4 & 93.0  \\
ru & 59.1 & 61.5 & 67.4 & 67.9 & 92.2 & 92.7 & 95.5 & 95.4  \\
sw & 68.4 & 71.1 & 74.9 & 72.5 & 94.7 & 95.6 & 96.7 &  97.2 \\
te & 81.3 & 75.2 & 84.6 & 83.4 & 97.6 & 98.0 & 99.2  &  99.0 \\
th & 75.0 & 75.2 & 80.2 & 78.6 & 98.2 & 98.0 & 98.9 &  98.7 \\
zh & 45.9 & 51.5 & 56.0 & 56.2 & 87.9 & 92.1 & 93.3 & 94.9  \\ \hline
Avg & 60.8 & 62.3 & 66.5 & 65.7 & 92.4 & 93.1 & 94.3 & 94.6  \\ \hline
\end{tabular}}
\caption{nDCG@10 and R@100 on the development set of the MIRACL dataset.}
\label{tab:app_full_miracl_ndcg}
\end{table*}

\section{Implementation Details}

\noindent
\textbf{Contrastive Pre-training Text Pairs }
In Table ~\ref{tab:pretrain_data},
to construct text pairs,
we utilize (section title, section passage) for Wikipedia,
(title, page content) for mC4~\citep{xue2021mt5},
(title, news content) for multilingual CCNews~\footnote{\url{https://commoncrawl.org/blog/news-dataset-available}},
translation pairs for NLLB~\citep{costa2022no},
(comment, response) for Reddit~\footnote{\url{https://www.reddit.com/}},
(title, abstract) and citation pairs for S2ORC~\citep{lo2020s2orc},
(question, answer) for Stackexchange~\footnote{\url{https://stackexchange.com/}},
(input prompt, response) for xP3~\citep{muennighoff2022crosslingual}.
For the miscellaneous SBERT data~\footnote{\url{https://huggingface.co/datasets/sentence-transformers/embedding-training-data}},
we include the following datasets:
SimpleWiki, WikiAnswers, AGNews, AltLex, AmazonQA, AmazonReview,
CNN/DailyMail, CodeSearchNet, Flickr30k, GooAQ, NPR, SearchQA,
SentenceCompression, Specter, WikiHow, XSum, and YahooAnswers.
\newline

\noindent
\textbf{Data Mixture for Supervised Fine-tuning }
It includes ELI5~\citep{fan2019eli5}(sample at $20\%$), HotpotQA~\citep{yang2018hotpotqa},
FEVER~\citep{Thorne2018FEVERAL}, MIRACL~\citep{Zhang2023MIRACLAM},
MSMARCO passage ranking and document ranking (sample at $20\%$)~\citep{Campos2016MSMA},
NQ~\citep{Karpukhin2020DensePR}, NLLB (sample at $100k$)~\citep{costa2022no},
NLI~\citep{Gao2021SimCSESC}, SQuAD~\citep{Karpukhin2020DensePR}, TriviaQA~\citep{Karpukhin2020DensePR},
Quora Duplicate Questions~\citep{quora-question-pairs}(sample at $10\%$), MrTyDi~\citep{zhang2021mr},
and DuReader~\citep{qiu2022dureader} datasets.

For the \emph{mE5-large-instruct} model,
we employ the new data mixture from ~\citet{wang2023improving}.
The main difference is the inclusion of synthetic data from GPT-4.
\newline

\noindent
\textbf{Training Hyperparameters }
The mE5$_\text{small}$, mE5$_\text{base}$ and mE5$_\text{large}$
are initialized from the multilingual MiniLM ~\citep{wang2021minilmv2},
\emph{xlm-roberta-base}~\citep{conneau2020unsupervised},
and \emph{xlm-roberta-large} respectively.
For contrastive pre-training,
the learning rate is set to \{$3, 2, 1$\}$\times10^{-4}$ for the \{small, base, large\} models.
For fine-tuning,
we use batch size $512$ and
learning rate \{$3, 2, 1$\}$\times10^{-5}$ for the \{small, base, large\} models.
All models are fine-tuned for $2$ epochs.
The \emph{mE5-large-instruct} model adopts the same hyperparameters as the mE5$_\text{large}$ large,
but is fine-tuned on the new data mixture by ~\citet{wang2023improving}.

\begin{table*}[ht]
\centering
\scalebox{0.88}{\begin{tabular}{lcccc}
\hline
Dataset & mE5$_\text{small}$ & mE5$_\text{base}$ & mE5$_\text{large}$ & mE5$_\text{large-instruct}$ \\ \hline
BIOSSES & 82.3 & 85.1 & 82.5 & 87.0 \\
SICK-R & 77.5 & 78.5 & 80.2 & 81.7 \\
STS12 & 76.6 & 76.7 & 80.0 & 82.6 \\
STS13 & 77.0 & 78.0 & 81.5 & 87.2 \\
STS14 & 75.5 & 76.6 & 77.7 & 85.0 \\
STS15 & 87.1 & 88.2 & 89.3 & 91.0 \\
STS16 & 83.6 & 84.3 & 85.8 & 87.3 \\
STS17 & 86.4 & 87.8 & 88.1 & 90.0 \\
STS22 & 60.9 & 61.8 & 63.1 & 67.6 \\
STSBenchmark & 84.0 & 85.6 & 87.3 & 88.4 \\
SummEval & 30.0 & 30.1 & 29.7 & 30.4 \\
SprintDuplicateQuestions & 92.2 & 93.0 & 93.1 & 91.2 \\
TwitterSemEval2015 & 70.8 & 72.2 & 75.3 & 80.3 \\
TwitterURLCorpus & 84.8 & 85.5 & 85.8 & 87.1 \\
AmazonCounterfactualClassification & 73.8 & 79.0 & 79.1 & 76.2 \\
AmazonPolarityClassification & 88.7 & 90.6 & 93.5 & 96.3 \\
AmazonReviewsClassification & 44.7 & 44.5 & 47.6 & 56.7 \\
Banking77Classification & 79.4 & 82.7 & 84.7 & 85.7 \\
EmotionClassification & 42.5 & 45.2 & 46.5 & 51.5 \\
ImdbClassification & 80.8 & 85.5 & 90.2 & 94.6 \\
MassiveIntentClassification & 70.3 & 72.1 & 73.8 & 77.1 \\
MassiveScenarioClassification & 74.5 & 77.1 & 77.5 & 80.5 \\
MTOPDomainClassification & 91.1 & 93.1 & 93.7 & 93.9 \\
MTOPIntentClassification & 71.1 & 75.3 & 77.9 & 82.5 \\
ToxicConversationsClassification & 69.4 & 69.8 & 71.3 & 71.1 \\
TweetSentimentExtractionClassification & 62.6 & 61.3 & 62.0 & 64.6 \\
AskUbuntuDupQuestions & 57.9 & 58.2 & 60.3 & 63.9 \\
MindSmallReranking & 30.3 & 31.0 & 31.4 & 33.1 \\
SciDocsRR & 78.1 & 80.7 & 82.0 & 85.9 \\
StackOverflowDupQuestions & 49.2 & 49.4 & 49.7 & 51.5 \\
ArxivClusteringP2P & 39.2 & 40.3 & 44.3 & 46.4 \\
ArxivClusteringS2S & 30.8 & 35.4 & 38.4 & 40.5 \\
BiorxivClusteringP2P & 35.8 & 35.0 & 35.3 & 40.9 \\
BiorxivClusteringS2S & 27.1 & 29.5 & 33.5 & 36.3 \\
MedrxivClusteringP2P & 30.9 & 28.9 & 31.5 & 36.9 \\
MedrxivClusteringS2S & 27.3 & 28.4 & 29.7 & 35.5 \\
RedditClustering & 39.1 & 42.4 & 46.5 & 56.6 \\
RedditClusteringP2P & 59.0 & 55.2 & 63.2 & 64.3 \\
StackExchangeClustering & 53.5 & 55.3 & 57.5 & 66.8 \\
StackExchangeClusteringP2P & 32.1 & 30.5 & 32.7 & 42.5 \\
TwentyNewsgroupsClustering & 33.2 & 36.0 & 38.9 & 51.3 \\
ArguAna & 39.1 & 44.2 & 54.4 & 58.4 \\
ClimateFEVER & 22.6 & 23.9 & 25.7 & 29.9 \\
CQADupstackAndroidRetrieval & 36.1 & 38.5 & 39.7 & 42.7 \\
DBPedia & 37.8 & 40.4 & 41.3 & 38.4 \\
FEVER & 75.3 & 79.4 & 82.8 & 78.0 \\
FiQA2018 & 33.3 & 38.2 & 43.8 & 47.7 \\
HotpotQA & 65.1 & 68.6 & 71.2 & 69.3 \\
MSMARCO & 41.0 & 42.3 & 43.7 & 40.4 \\
NFCorpus & 31.0 & 32.5 & 34.0 & 35.5 \\
NQ & 56.3 & 60.0 & 64.1 & 57.8 \\
QuoraRetrieval & 86.9 & 87.7 & 88.2 & 89.2 \\
SCIDOCS & 13.9 & 17.2 & 17.5 & 18.7 \\
SciFact & 67.7 & 69.3 & 70.4 & 71.8 \\
Touche2020 & 21.2 & 21.4 & 23.4 & 27.2 \\
TRECCOVID & 72.6 & 69.8 & 71.3 & 82.0 \\ \hline
Average & 57.9 & 59.4 & 61.5 & \textbf{64.4} \\ \hline
\end{tabular}}
\caption{Results for each dataset in the MTEB benchmark.
The evaluation metrics are available in the original paper~\citep{muennighoff2023mteb}.}
\label{tab:app_full_results}
\end{table*}

\end{document}